\documentclass{article}

\usepackage{times}
\usepackage{graphicx} 
\usepackage{subfigure} 

\usepackage{natbib}

\usepackage{algorithm}
\usepackage{algorithmic}
\usepackage{multirow}

\usepackage{hyperref}

\usepackage[accepted]{data4good2016}

\icmltitlerunning{Boosting International Understanding and Transparency in News Coverage}

\begin{document}

\twocolumn[
\icmltitle{Machine Learning meets Data-Driven Journalism:\\ Boosting International Understanding and Transparency in News Coverage}


\icmlauthor{Elena Erdmann$^{1}$, Karin Boczek$^{2}$, Lars Koppers$^{3}$, Gerret von Nordheim$^{2}$, Christian P{\"o}litz$^{1}$, Alejandro Molina$^{1}$, Katharina Morik$^{1}$, Henrik M{\"u}ller$^{2}$, J{\"o}rg Rahnenf{\"u}hrer$^{3}$, Kristian Kersting$^{1}$}{fname.lname@tu-dortmund.de}
\icmladdress{$^{1}$Computer Science Department, $^{2}$Institute for Journalism, $^{3}$Department of Statistics, TU Dortmund University, Germany}

\icmlkeywords{text mining, data-driven journalism, machine learning, ICML}

\vskip 0.3in
]


\begin{abstract} 
Migration crisis, climate change or tax havens: Global challenges need global solutions. But agreeing on a joint approach is difficult without a common ground for discussion. Public spheres are highly segmented because news are mainly produced and received on a national level. Gaining a 
global view on international debates about important issues is hindered by the enormous quantity of news and by language barriers. Media analysis usually focuses only on qualitative research. In this position statement, we argue that it is imperative to pool methods from machine learning, journalism studies and statistics to help bridging the segmented data of the international public sphere, using the Transatlantic Trade and Investment Partnership (TTIP) as a case study.
\end{abstract}

\section{The need for cross-national analysis}
The recently published news on the Panama Papers leak demonstrates firstly that tax fraud is an international phenomenon and secondly how cross-national cooperation can be beneficial to investigating and reporting. Admittedly, this is an exceptional case. Global events are still "primarily covered in accordance with the traditional national outlook, i.e. national domestications and the 'domestic vs. foreign news” logic'" \citep[847]{berglez_what_2008}. A global public sphere to address globally relevant issues has not been established yet and national biases impede possible international approaches. This way "the global sociopolitical order becomes defined by the realpolitik of nation-states that cling to the illusion of sovereignty despite the realities wrought by globalization" \citep[80]{castells_new_2008}. Reciprocal knowledge about controversial issues across national borders is necessary to provide common ground for fruitful global discussions and proposals.      

In this position statement, we provide evidence that joining forces improves media transparency on a global scale: Combining machine learning with statistics and journalism studies contributes to bridging the segmented data of the international public sphere. 
Following an interdisciplinary approach we tackle the question of how methods from machine learning help to deepen our understanding of the discussion on cross-national issues. LDA, @TM, PDNs and word2vec are used to enhance transparency on international media coverage: Range, amount and framing of issues can be compared with fewer translation efforts. Differences in perception become obvious and evaluation divides can be interpreted. This will be demonstrated by an analysis of the coverage on the controversial Transatlantic Trade and Investment Partnership (TTIP) between the United States of America (U.S.) and the European Union (E.U.).      

\begin{figure*} 
\centering
\subfigure[Comparison of number of articles containing 'TTIP' in New York Times and S\"uddeutsche Zeitung.\label{keywordsearch}]{
\includegraphics[scale=0.32]{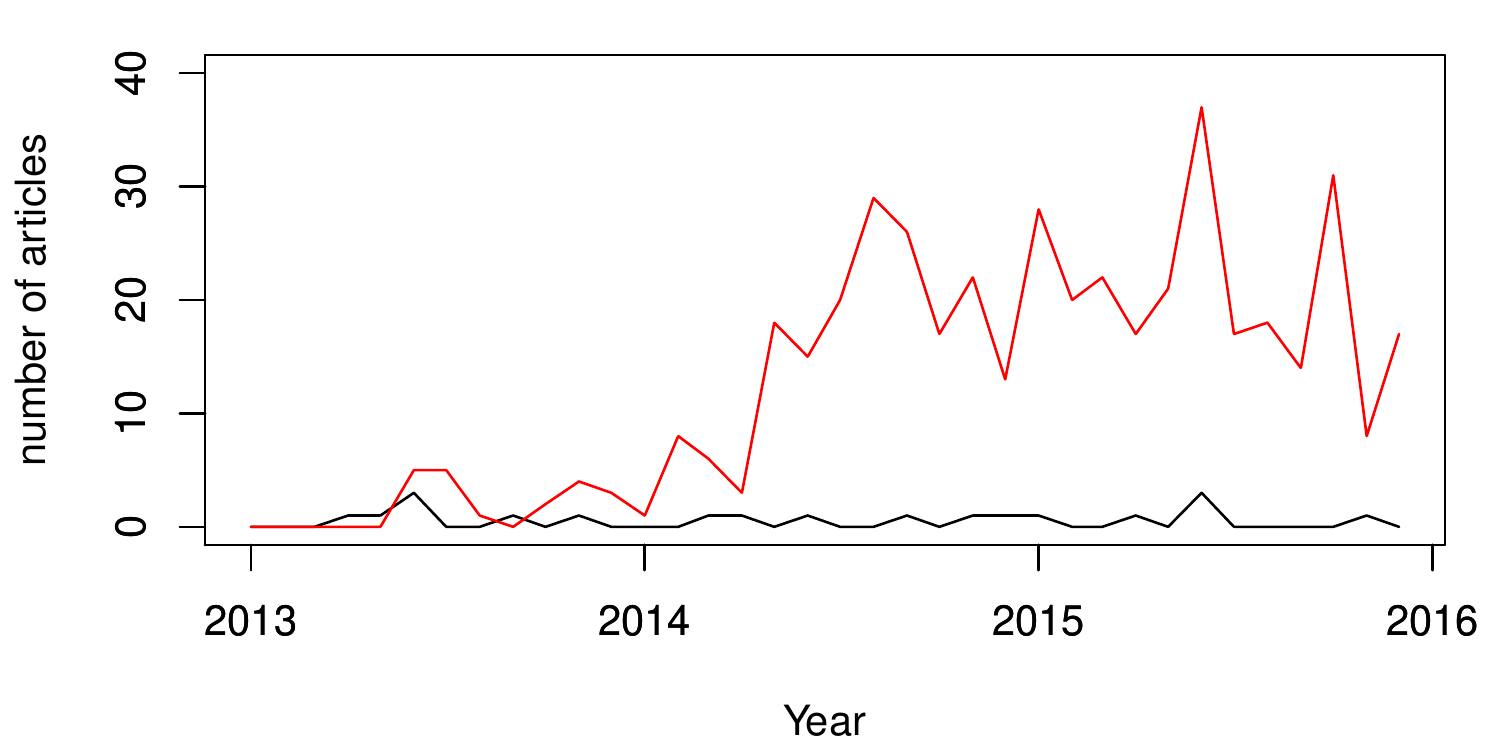}
}
\quad
\subfigure[PDN created from the articles containing 'TTIP' in S\"uddeutsche Zeitung. (German words were translated to English)\label{pdnsz}]{
\includegraphics[trim = 90mm 20mm 0mm 80mm,clip,scale=0.29]{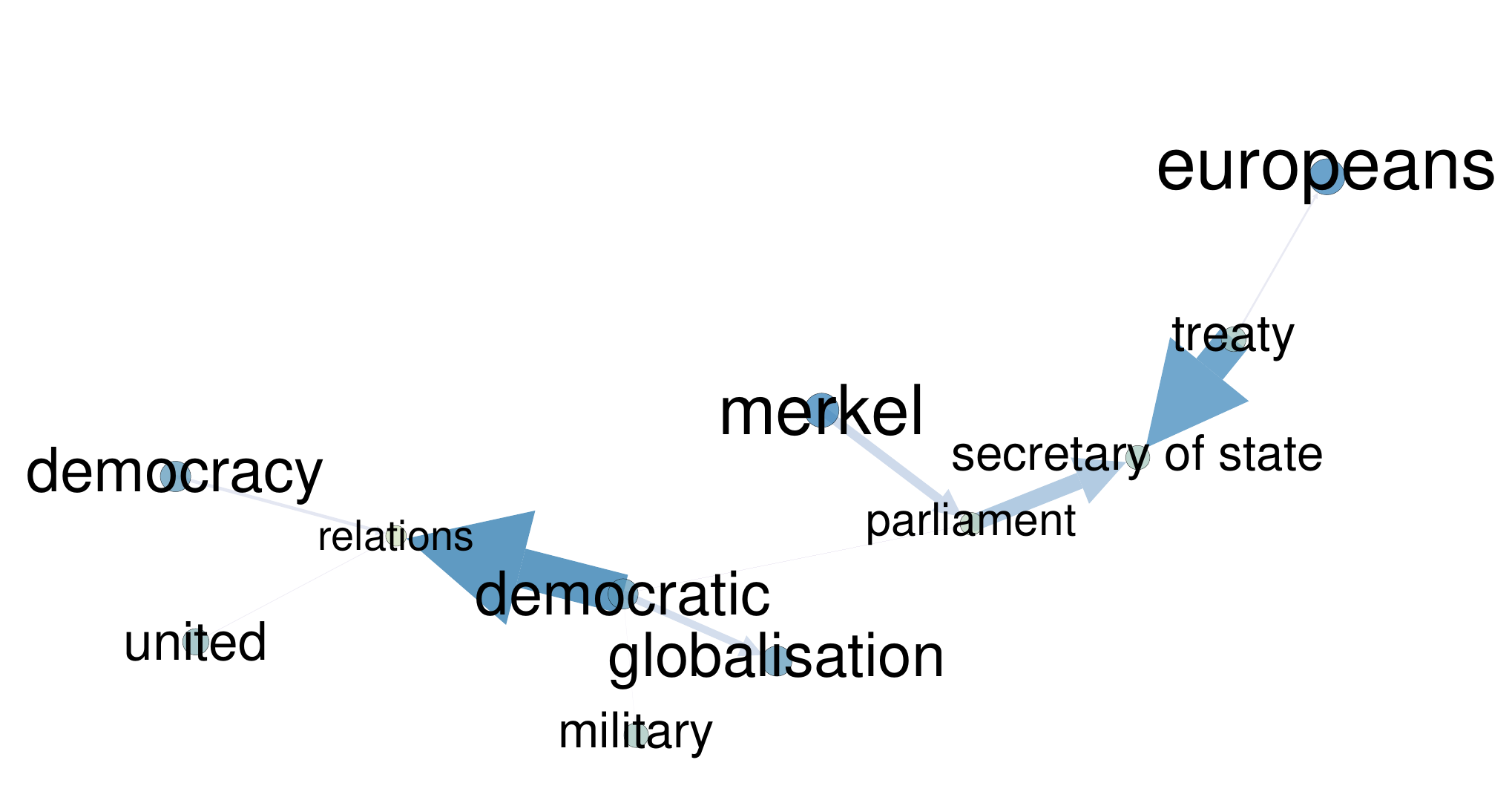}
}
\quad
\subfigure[Attentional curves capture the development of topics in news articles over time, here illustrated for the war on Ukraine.\label{ukraineGompertz}]{
\includegraphics[scale=0.19]{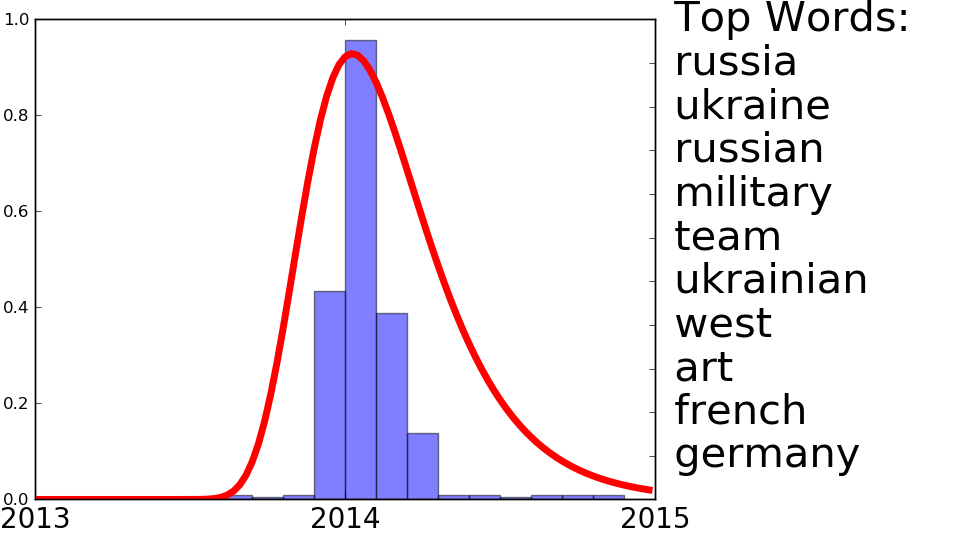}
}
\caption{Boosting international understanding and transparency in news coverage using machine learning and data-driven journalism.}
\end{figure*}

TTIP was designed to facilitate trade between the U.S. and the E.U. 
However, TTIP's actual impact on economies and societies has been discussed controversially in both the U.S. and Europe. Media perception has differed in many aspects. The comparison of a U.S. newspaper (New York Times) and a German newspaper (S\"uddeutsche Zeitung) reveals that TTIP is more hotly debated in Germany than in the U.S., see Fig.~\ref{keywordsearch}.
The New York Times highlighted the need of bank regulations and the threat that exporting nations pose to local markets. Whereas, S\"uddeutsche Zeitung focused largely on consumer protection. On the one hand TTIP was criticized for its implications on environmental and food standards, on the other for the negotiation proceedings that were characterized by democratic deficits and insufficient transparency. Intricate questions derive from the simple comparison of word frequencies: Why does the range of reported arguments differ so considerably? Is the German media reluctant to the Trade Partnership in general? And what are the reasons for the German obsession with the 'chlorinated chicken' as evidenced in the significant number of these words in articles broaching the TTIP issue? 

The TTIP example illustrates that media coverage can largely differ across nations. Public discussion is still strongly influenced by national media. Multiple languages add to the difficulties to frame a common global perspective. However, in a globalized world, crisis and political decisions have become far too complex to be dealt with on a national level only. Combining Machine Learning and data-driven journalism
enables researchers to investigate large corpora of texts to reveal national patterns of argumentation, which in turn can promote international understanding.

\section{ML meets Data-Driven Journalism}



There is an arms race to `deeply' understand text data, and consequently
a range of different techniques has been developed for media analysis. However, 
when using them for data-driven journalism, e.g. to gain a deeper understanding of the news reception of important political and societal issues, there are also challenges. 


Just to name few of the recent ML techniques, DeepDive \cite{niu12} aims to extract structured data from texts, Metro Maps \cite{sha12} extract easy to understand networks of news stories, and word2vec \cite{mik13} computes Euclidean embeddings of words. It is trained on a corpus of documents and transforms each word into a vector by calculating word correlations. Similarity measures can be applied to the resulting vectors. Particularly, word2vec can be used to compute those words that are most likely to occur in the same context as a given word. It thus has the potential to reveal which words are linked closest to a given issue and hence provides a semantically enriched alternative to classical keyword searches, which is well used by journalists. Finally, 
topic models, have been used successfully in many scenarios, in particular to model discourses. Most prominent among them is Latent Dirichlet Allocation (LDA) \cite{blei03} that characterizes each topic as a list of words and their respective probabilities to appear in the topic.
Topics over Time (TOT) \cite{wan06} follows this paradigm, but introduces a temporal component. In TOT, each document has a timestamp and the probability of a topic grows and declines over time. Thus, TOT can be employed to analyze trends in news. 

Due to this rich machine learning toolbox for analyzing news articles, it is tempting to put a stack of news articles on a data journalist's desk saying `Enjoy'. Unfortunately, data-driven journalism is not that simple.

Reconsider topic models, the main focus of the present paper. TOT does not model attention of the crowd in a physically plausible way.
Triggered by models from communication studies \cite{kolb05} and the observation that the Shifted Gompertz distribution models attentional curves \cite{bau14}, we developed a novel Attentional Topic Model (@TM) \cite{poe16}. It captures well the growth and decline of the popularity
of topics in a physically plausible way.

Moreover, multinomial word distributions, such as in LDA and TOT capture the most common words used in each topic. However, they often fail to give a deeper understanding of topics required when investigating media discourse. That is why APMs \cite{ino14}, which discover word dependencies in each topic, have been introduced;
essentially, they encode topics as weighted undirected graphs. Often, however, word dependencies are asymmetric.
If the word 'treaty' appears in a text, it is very likely that
the text will refer to the museum's 'secretary of state',
too. The phrase 'secretary of state', on the other hand,
is a very general term and can be used in many different contexts. Thus, it does not make the word
'treaty' per se more likely. In \cite{erd16}, we therefore extended APMs to directed dependencies using Poisson Dependency Networks \cite{had15}. Moreover, longer chains of directed dependencies may provide interesting clues to understand a topic.

Finally, topic models have been traditionally evaluated using intrinsic measurements such as the likelihood and the perplexity of topics \cite{wal09}. As these measurements do not necessarily correspond to human judgment \cite{cha09}, we pool together the talents of journalists, machine learners and statisticians to obtain a better understanding of what makes a good topic. If we use topic models to create subcorpora e.g.\ for content analysis we have to ensure that the subcorpora are at least as good as the ones from other methods like keyword searches. The gold standard is the evaluation based on human judgment \cite{stryker_validation_2006}. The use of statistical methods helps to reduce the time requirement for human coders to come to a significant statement about the quality of a subcorpus. Moreover, our interdisciplinary research led to several interesting observations about the quality of topics: While researchers from a mathematical background tended to focus on topics linked to large quantities of documents, journalists oftentimes preferred those topics that were created from only few meaningful documents. Likewise words like 'can', 'need' and 'do', that were considered stopwords by machine learners, really caught the journalist's attention.

We believe that these small and seemingly insignificant notices can help to improve the application and lead a way to new computational models. Can new topic models be developed to cater better to the specific needs of journalists? Are there different approaches to gain deeper insight into each topic? We will illustrate this using the case of TTIP.

\section{Towards an International View on TTIP}

TTIP affects millions of people living in the U.S.\ and the E.U.\ and its negotiations have been controversial. However, content analysis of newspapers indicates that the issue is of diverging national importance.
Both compared newspapers are high-circulation dailies from metropolises that exhibit a rather liberal orientation. Despite these similarities, coverage on TTIP varies significantly (see Fig.~\ref{keywordsearch}). 

It is notable that coverage on TTIP increases considerably from 2014 in S\"uddeutsche Zeitung (SZ), whereas the number of articles in the New York Times (NYT) remains unaltered. In order to shed light on this disparity the general coverage on the U.S.A. and Europe, respectively, was analyzed. In SZ the sub-corpus with all articles including the pattern of the letters \verb+usa+ contained 59.637 articles (36 per cent of the corpus) whereas the \verb+europe+-corpus in NYT contained  34.177 (11 per cent of the corpus).
\begin{figure}[b]
\centering
\caption{Topics in NYT found by LDA and labeled by journalists.\label{lda}}
\includegraphics[scale=0.2]{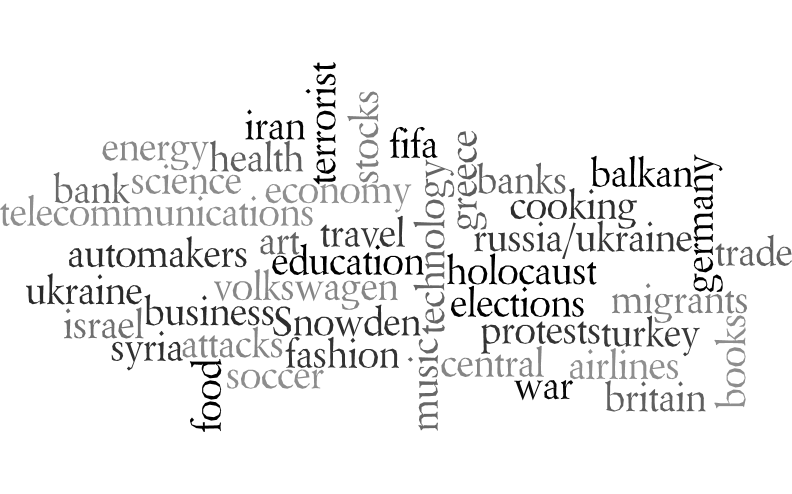}
\end{figure}

{\bf (Topic Models)} LDA was used to find 100 topics in each sub-corpus. The topics were labeled by journalists using top words and top articles. The topics in NYT are illustrated in Fig.~\ref{lda}. A glance at the results shows distinctly the different perspectives on TTIP in the public spheres. In SZ TTIP appears on the top word list of a topic  covering articles on policies of the European Commission along with conjoined words like customs, arbitration and investor protection. In NYT TTIP is not among the top words of the European Commission topic which is dominated by stakeholders dealing with financial issues around the euro crisis. Interpreting the LDA topics further TTIP plays a less significant role in the U.S.-European relations represented by the LDA topics which are mainly various international conflicts, art, sports and general economy related topics.

{\bf (Directed word dependencies)} A PDN \cite{had15} trained on the articles of S\"uddeutsche Zeitung shows that the constitution of the E.U.\ as a politico-economic union of 28 states places the question of parliamentary participation in the foreground (see Fig.~\ref{pdnsz}). In the U.S., this question does not arise.

{\bf (Attentional Topic Models)} When analyzing the discourse on Europe in NYT through Attentional Topic Models \cite{poe16}, we found no attentional topic corresponding to the TTIP. Instead, the discussion focused on different topics such as the war in the Ukraine (see Fig.~\ref{ukraineGompertz}).

{\bf (word2vec)} Analyzing word2vec results highlights diverse reciprocal perception: U.S. and German newspaper both coincide covering Germany mainly considering the recent migration. However, on the coverage on the U.S. SZ and NYT drift apart: In the SZ the importance of U.S. as an economic partner is demonstrated, whereas the NYT covers the U.S. in a broader range of topics including several sports (see Table \ref{w2v}). Comparing the use of TTIP indicates that the NYT uses more matter-of-fact words in connection with TTIP while SZ seems to include more commenting and evaluating words including 'chlorinated chicken'. Word2vec solves the mystery of the 'chlorinated chicken': free trade agreement, genetically modified food and genetically modified corn are among the most similar words. For German TTIP opponents chicken meat disinfected with chlorine has become a symbol for lowering food safety standards and the disadvantages of TTIP in general.



 Applying machine learning to understand the cross-nationally diverse discourse on TTIP highlights the benefit which can be derived from an interdisciplinary approach. Yet, this interdisciplinary approach is still in its infancy.

\begin{table}[t]
\begin{center}
\begin{tabular}{p{1.2cm}p{6cm}}
    \hline
        \multirow{2}{*}{USA}& \tiny{NYT: santander consumer, served chairman, basketball, womens hockey, oracle team, senior vice, kan, goodgame, divac, columbus ohio}\\
    &\tiny{SZ: kanada mexico, united states, china hongkong, largest market, most important trade partner, embargo, pacific states, china russia, great britain france, usa kanada}\\
    \hline
        \multirow{2}{*}{Germany}& \tiny{NYT: germanys, europe, asylum seekers, migrants, german, plan distribute, migrants entered, accept migrants, hungary closed, human flow}\\
    &\tiny{SZ: europe, countries, immigrant, kanada australia, prospects of remaining, immigrants, arriving refugees, many refugees, most refugees, european countries}\\
    \hline
        \multirow{2}{*}{TTIP}& \tiny{NYT: transatlantic trade, investment partnership, trade ministers, mr froman, trade agreement, trade negotiations, trade negotiators, trade talks, euus, trade commissioner}\\
    &\tiny{SZ: ceta, investor protection, free trade agreement, free trade agreement ttip, trade agreement, investment protection, eu kanada, transatlantic free trade agreement, chlorine chicken, free trade}\\
    \hline
\end{tabular}
\end{center}
\caption{Most similar words to 'USA', 'Germany' and 'TTIP' in NYT and SZ (translated from German) according to word2Vec.\label{w2v}}
\end{table}

\section{Lesson Learned: Bridging Fields}

The absence of a common public sphere has already been constituted as an enduring obstacle to further political and economic integration in Europe \cite{habermas_crisis_2014,jones_failing_2015,vossing_transforming_2015,risse_european_2015}, a region where the difficulty of understanding between people speaking different languages becomes evident in spite of small distances. 
Agents in politics and business face a confusing multitude of partly conflicting national discourses. Therefore, finding common solutions in a democratic context is hindered. The defiance of finding common ground for discussion becomes even more challenging if international understanding is volitional.    
To amend the development of a global public sphere discussing and approaching international challenges it is imperative that computer scientists, information scientists, and experts in communication studies pool their talents and knowledge to help find efficient and effective ways of managing the news  sources available. Research in this new field is necessarily interdisciplinary since developing new methods and applying established ones should eventually lead to instruments that enable not only researches and experienced data journalists but also practitioners in the media, in politics and business to compare debates internationally. 

So far using machine learning methods for content analysis is still uncommon in communication studies and best practices for algorithmic text analysis (ATA) are still being negotiated. In the TTIP case they were used as part of a hybrid approach "that combines computational and manual methods throughout the process . . . [to] retain the strengths of traditional content analysis while maximizing the accuracy, efficiency, and largescale capacity of algorithms for examining Big Data." \cite{lewis_content_2013} Following this approach, patterns that so far have been hidden can be made visible. Transparency will be achieved on the prevailing debates, showing how they evolve and relate to existing narratives, identifying national frames and agenda setters and showing divergences and convergences across national debates. 
The focus of research should be on interlocking of long-term discourse patterns with current issues. Which arguments and frames have dominated the debate on the refugee crisis? Why is the opposition against TTIP so strong in the German speaking countries of Germany, Austria and Luxembourg while others embrace the deal? Why have Germany and France differed so fundamentally on how to handle the Greek debt crisis? Which persons and institutions are dominating the debates in the respective countries? In which areas do new topics or new frames emerge?	

Our TTIP study also motivates to revisit a considerable number of important theories of communication studies. The mechanisms of Agenda-setting \cite{bennett_toward_2006} and issue attention cycles \cite{downs_up_1972} can be visualized using clustering models in an entirely new dimension; the most important agents of the public discourse \cite{habermas_structural_1991} can be analyzed with named-entity-recognition and network-visualizations, framing of news \cite{entman_framing:_1993} can be illustrated using sentiment-analysis. In a nutshell, the potential of machine learning for analyzing international communication and discourses is high. However, the potential can only be achieved if new methods are developed and made available as easy-to-use applications. 

Overall, applying machine learning to broaden insight in international news coverage opens up fundamentally new intellectual territory with great potential to advance the state of the art of computer science and related disciplines and to provide unique societal benefits. Measures to achieve this potential involve intense interdisciplinary collaboration and the mutual objective to develop methods being easily usable for everyone interested in profound international understanding.

\section*{Acknowledgment}
This work was supported by the DFG Collaborative Research Center SFB 876 project A6 and A1 and the Dortmund Center for Media Analysis (DoCMA).

\bibliography{example_paper}
\bibliographystyle{icml2016}

\end{document}